\pdfoutput=1

\documentclass[11pt]{article}

\usepackage[]{acl}

\usepackage{times}
\usepackage{latexsym}

\usepackage[T1]{fontenc}

\usepackage[utf8]{inputenc}

\usepackage{microtype}

\usepackage{inconsolata}

\newcommand{\MODEL}{\mbox{\textsc{Coin}}\xspace}
\newcommand{\MODELFULL}{\underline{Co}ntrastive \underline{In}struction Tuning (\mbox{\textsc{Coin})}}

\usepackage{refstyle}
\usepackage{amsmath}
\usepackage{cleveref}

\usepackage{booktabs}
\usepackage{multirow} %
\usepackage{soul}%
\usepackage{tabularx}

\usepackage{xspace}
\usepackage{pifont}

\usepackage{graphicx}

\usepackage{makecell}

\crefformat{section}{\S#2#1#3}
\crefformat{subsection}{\S#2#1#3}
\crefformat{subsubsection}{\S#2#1#3}
\crefrangeformat{section}{\S#3#1#4 to~\S#5#2#6}
\crefmultiformat{section}{\S#2#1#3}{ and~\S#2#1#3}{, #2#1#3}{ and~#2#1#3}
\Crefformat{figure}{#2Fig.~#1#3}
\Crefmultiformat{figure}{Figs.~#2#1#3}{ and~#2#1#3}{, #2#1#3}{ and~#2#1#3}
\Crefformat{table}{#2Tab.~#1#3}
\Crefmultiformat{table}{Tabs.~#2#1#3}{ and~#2#1#3}{, #2#1#3}{ and~#2#1#3}
\Crefformat{appendix}{#2Appx.~\S#1#3}
\crefformat{algorithm}{Alg.~#2#1#3}
\Crefformat{equation}{#2Eq.~#1#3}

\newcommand{\stitle}[1]{\vspace{1ex} \noindent{\bf #1.}}

\title{Contrastive Instruction Tuning}

\author{
Tianyi Lorena Yan$^\diamondsuit$, Fei Wang$^\diamondsuit$, James Y. Huang$^\diamondsuit$, Wenxuan Zhou$^\diamondsuit$, Fan Yin$^\spadesuit$ \\
\textbf{Aram Galstyan$^\diamondsuit$, Wenpeng Yin$^\heartsuit$, Muhao Chen$^\clubsuit$}
 \\
  $^\diamondsuit$University of Southern California~~~~$^\spadesuit$University of California, Los Angeles \\
  $^\heartsuit$The Pennsylvania State University~~~~$^\clubsuit$University of California, Davis \\
   \texttt{\{tianyiy, fwang598, huangjam, zhouwenx\}@usc.edu}~~~~\texttt{fanyin20@cs.ucla.edu} \\ 
    \texttt{galstyan@isi.edu}~~~~\texttt{wenpeng@psu.edu}~~~~\texttt{muhchen@ucdavis.edu}
}

\begin{document}
\maketitle

\begin{abstract}

Instruction tuning has been used as a promising approach to improve the performance of large language models (LLMs) on unseen tasks. However, current LLMs exhibit limited robustness to unseen instructions, generating inconsistent outputs when the same instruction is phrased with slightly varied forms or language styles. This behavior indicates LLMs' lack of robustness to textual variations and generalizability to unseen instructions, potentially leading to trustworthiness issues. Accordingly, we propose \MODELFULL, which maximizes the similarity between the hidden representations of semantically equivalent instruction-instance pairs while minimizing the similarity between semantically different ones. To facilitate this approach, we augment the existing FLAN collection by paraphrasing task instructions. Experiments on the PromptBench benchmark show that \MODEL consistently improves LLMs' robustness to unseen instructions with variations across character, word, sentence, and semantic levels by an average of $+2.5\%$ in accuracy.\footnote{Code is available at \url{https://github.com/luka-group/CoIN}.}

\end{abstract}

\section{Introduction}
\label{sec/introduction}
Instruction tuning has emerged to be an essential training paradigm of large language models (LLMs; \citealt{wei_finetuned_2022, sanh_multitask_2022, mishra_cross-task_2022}). By training models on various pairs of task instructions and instances, instruction tuning has been widely adopted in LLMs, such as TK-Instruct \cite{wang_super-naturalinstructions_2022}, InstructGPT \cite{ouyang_training_2022}, FLAN-T5 \cite{wei_finetuned_2022}, and Alpaca \cite{taori_stanford_2023}, allowing them to %
follow various human instructions and fulfill user intents %
\citep{wang_super-naturalinstructions_2022, zhang_instruction_2023}.

\begin{figure}
    \centering
    \includegraphics[width=1\linewidth]{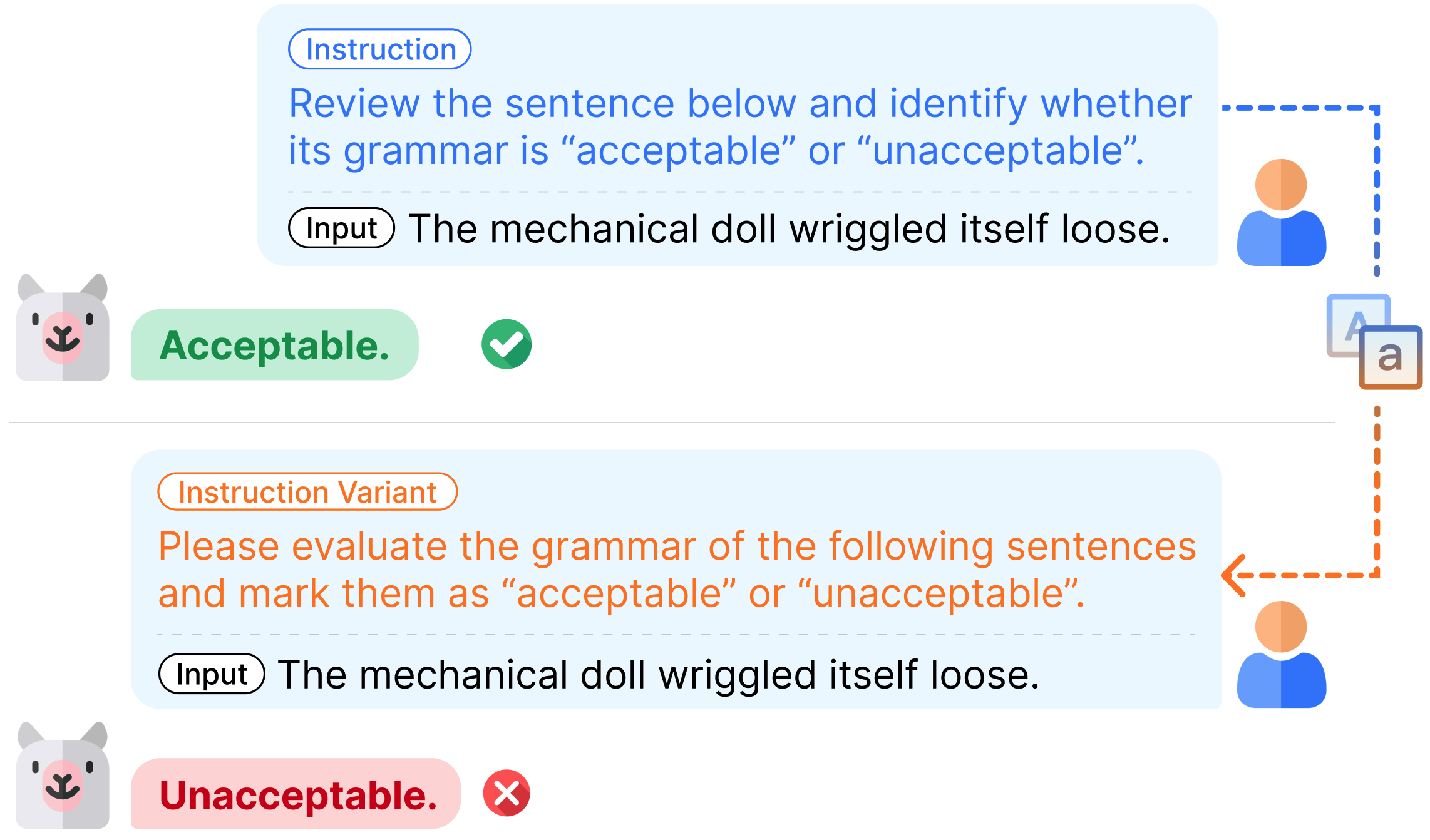}
    \caption{An example from CoLA \cite{warstadt_neural_2019} shows that current LLMs like Alpaca may generate entirely different responses when presented with semantically equivalent but textually different instructions. 
    }
    \label{fig:figure_1_different_answer}
\end{figure}

Despite these advancements, current instruction-tuned LLMs are not robust to instruction variations. Their performance may vary significantly when one re-formulates an instruction with different forms or language styles \citep{zhu_promptbench_2023, liu_robustness_2023}. %
While optimal instructions for specific user intents may exist, there is no guarantee that user-crafted instructions will precisely match them. Indeed, user-crafted instructions often contain variations that can cause drop in LLMs' performance, such as unintended minor mistakes (e.g., a typo; \citealt{wang_adversarial_2021,wang_robustness_2023}), different expression preferences (e.g., choice of synonyms or paraphrases; \citealt{gu_robustness_2023,zhu_promptbench_2023}), inefficient descriptions \cite{sun_evaluating_2023}, and varying formats \cite{liang_exploring_2023}.
As shown in \Cref{fig:figure_1_different_answer}, given different instructions of the same intent, an instruction-tuned LLM like Alpaca can generate entirely different responses, some of which can lead to wrong answers.  %
LLMs' current lack of robustness to instruction variations severely limits their real-world applications. However, prior instruction tuning methods mainly focus on aligning the desired output for a given instruction-input pair and do not explicitly address models' robustness against variations in instructions \cite{ouyang_training_2022, wei_finetuned_2022, zhang_instruction_2023, longpre_flan_2023}.

In this paper, we propose \MODELFULL, an instruction tuning method that leverages contrastive learning to %
align the hidden representations of instruction-instance pairs that are semantically equivalent but textually different and to differentiate those that are semantically distinct. Given the same input and output instance, we pair each instruction with its perturbed versions as positive samples. Observed that the hidden representations of data from different tasks already have low cosine similarity with each other \cite{liu_how_2023}, we use the same instruction paired with different instance input and output as hard negative samples (refer to \Cref{section/pos_neg_selection} for more details). %
Intuitively, %
by recognizing that the same instruction with different formulations can have the same meaning, %
the model can generate more consistent answers given different instructions of the same intent and become more robust to variations in language expressions. At the same time, negative samples encourage the model to understand that an instruction can lead to different outcomes in different contexts, facilitating the model to distinguish inputs with different user intents.

We assess LLMs' robustness on the PromptBench benchmark~\cite{zhu_promptbench_2023}, which introduces variations to instructions of a diverse set of tasks at character, word, sentence, and semantic levels. Experiments on the benchmark show that \MODEL significantly improves task performance and reduces response variation of Alpaca on \emph{unseen instructions} %
with variations at all four levels, achieving an average accuracy improvement of +2.5\% compared with continual instruction tuning on the same dataset.

Our contributions are three-fold. First, we propose a contrastive instruction tuning method, \MODEL, to enhance LLMs' robustness to semantic-invariant instruction variations. 
Second, experiments on PromptBench demonstrate the effectiveness of \MODEL in handling semantically equivalent instructions that differ at the character, word, sentence, and semantic levels. 
Third, to facilitate the proposed approach, we augmented the FLAN collection, a widely used instruction tuning dataset, with contrastive instructions. We will release the augmented dataset consisting of 52k entries and 104k instructions to support future work in this direction.

\section{Related Work}
\label{sec/related_work}
In this section, we provide a brief summary of three highly related topics.

\stitle{Instruction Tuning and Generalizability}
Instruction tuning has emerged to be one of the pivotal techniques for enhancing the generalizability of LLMs \citep{sanh_multitask_2022, zhang_instruction_2023, ouyang_training_2022}. This capability is crucial for LLMs, as it determines models' performance when encountering new data. The efficacy of instruction tuning has become more evident when the number of tasks scales up \cite{xu_zeroprompt_2022}. Consequently, many recent studies have been focusing on fine-tuning LLMs with a wide range of tasks. For instance, large-scale datasets that encompass numerous NLP tasks and multiple data sources have been curated for effectively enhancing LLMs' zero-shot generalizability \citep{sanh_multitask_2022, wei_finetuned_2022, niu_learning_2023, kung_active_2023, wang2023far}. %
Despite performance gained on unseen tasks, LLMs fine-tuned with large-scale instruction datasets remain vulnerable to how the same instruction is expressed differently \citep{wang_adversarial_2021, zhu_promptbench_2023, liu_robustness_2023, sun_evaluating_2023, liang_exploring_2023}. This limitation motivates us to enhance LLMs' robustness to instruction variations in this work.

\stitle{Robustness of Instruction-Tuned LLMs}
With the increasing reliance on LLMs, recent works have focused on having a comprehensive understanding of the robustness of instruction-tuned language models. \citet{zhu_promptbench_2023}, \citet{gu_robustness_2023}, and \citet{wang_robustness_2023} add perturbations to instructions %
across multiple levels (character, word, sentence, etc.) and show that current LLMs are not robust to the introduced perturbations. LLMs' performance can also be degraded when presented with unobserved, paraphrased versions of task instructions \cite{sun_evaluating_2023}. Furthermore, inconsistency in format and style in instruction expressions, such as placing instructions before, in between, or after the input instances, can decrease models' performance \cite{liang_exploring_2023}. While evaluating and analyzing LLMs' robustness has garnered more attention, enhancing the models' robustness, particularly against varied instructions of the same task, is an underexplored problem. Our work is dedicated to addressing this gap. 

\stitle{Contrastive Learning}
Contrastive learning, a self-supervised technique that involves training a model to contrast between positive and negative pairs of data points, has rapidly evolved and been adapted in NLP tasks, such as sentence embedding \cite{gao_simcse_2022}, summarization \cite{liu_brio_2022}, named entity recognition \cite{layegh_contrastner_2023}, and logical reasoning \cite{bao_enhancing_2023}. Within the context of instruction tuning, contrastive learning has been used with prefix-training to enhance the controllability towards desired attributes of LLMs \cite{qian_controllable_2022}. However, the focus of the work remains on steering the generated outputs towards an attribute (such as being sentimentally positive) that is assumed to be known but is difficult to be specified given the diversity of tasks that LLMs may handle, and it does not explicitly tackle the challenge of LLMs' robustness against variations in instruction expressions. Inspired by the observation that contrastive learning is suitable for aligning semantically related sentences \cite{gao_simcse_2022}, we encourage LLMs to learn the semantic invariance of varied instructions for the same task and aim to address LLMs' imperfect robustness at all four levels: character, word, sentence, and semantic.

\section{Contrastive Instruction Tuning}
In this section, we first provide a formal definition of contrastive instruction tuning (\Cref{sec/method/problem}). Then, we introduce contrastive sample selection (\Cref{section/pos_neg_selection}) and the contrastive loss (\Cref{sec/method/contrastive}) in our method \MODEL.

\subsection{Overview}
\label{sec/method/problem}
\begin{figure*}
    \centering
    \includegraphics[width=1\linewidth]{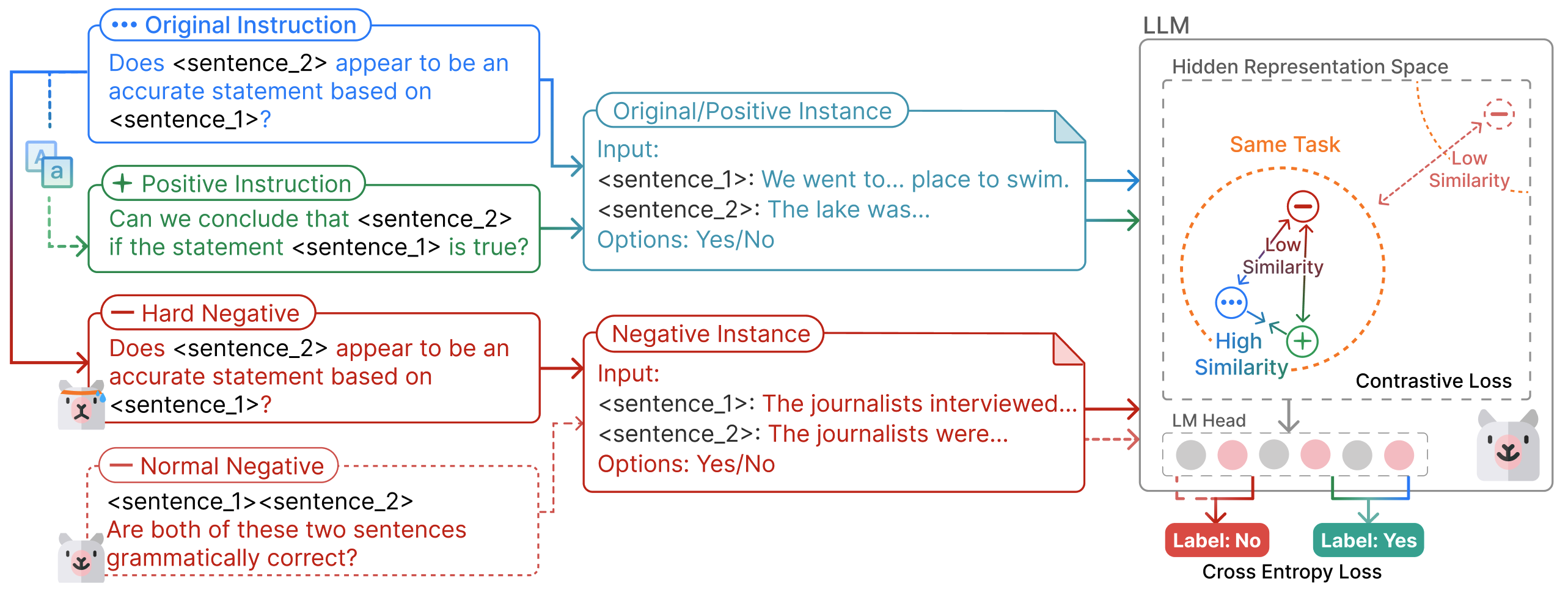}
    \caption{Illustration of \MODEL. A paraphrased instruction is used as the positive sample (green) given the same instance input and output. An instruction paired with different instance input and output is used as the negative sample (red). Cosine similarity between the hidden representations of original and paraphrased instruction-instance pairs is encouraged to be high, and vice versa for the paired negative samples. As we observe that the cosine similarity between the hidden representations of data from different tasks is already low \cite{liu_how_2023}, we use the same instruction paired with different instance input and output as hard negative samples to provide more informative training signals.}
    \label{fig:overview}
\end{figure*}
Assume that we have a (autoregressive) language model $\mathcal{M}$ and a dataset $\mathcal{D} = \{(I_i, x_i, y_i)\}_{i=1}^N$, in which $I_i$ denotes the task instruction, $x_i$ is the instance input, and $y_i$ is the desired output. For each original entry, we create a semantically equivalent entry $(I_i^+, x_i^+, y_i^+)$, where $x_i^+ = x_i$ and $y_i^+ = y_i$. $I_i^+$ is constructed by adding textual perturbations to the original instruction while ensuring the underlying semantic meaning remains the same. %
Our goal is to learn a model $\mathcal{M}$ such that its hidden representations of semantically equivalent instruction-instance pairs, denoted as $h_\mathcal{M}(I_i, x_i, y_i)$ and $h_\mathcal{M}(I_i^+, x_i^+, y_i^+)$, are close to each other in $\mathcal{M}$'s hidden representation space, thereby enhancing its robustness against instruction variations.

As explored by many previous studies, instruction-tuning with large-scale datasets mainly focuses on aligning the desired output for a given instruction-instance pair from various tasks \citep{sanh_multitask_2022, longpre_flan_2023, wei_finetuned_2022}. However, current LLMs exhibit a lack of robustness when facing the same instruction expressed in different forms \citep{sun_evaluating_2023, zhu_promptbench_2023, liang_exploring_2023}, causing LLMs to be unreliable when being deployed in the real world.
To mitigate this limitation, our method \MODEL further leverages contrastive learning to maximize the similarity between hidden representations of semantically equivalent instruction-instance pairs. This approach enhances models' robustness and consistency to variations in instruction expressions.

\subsection{Contrastive Data Selection}
\label{section/pos_neg_selection}
Selecting effective positive and negative samples for each instruction is critical to contrastive learning. In \MODEL, we construct positive samples by varying the phrasing or template structure of original instructions, ensuring that the positive samples still share the same input and output with the original instance. This approach helps the model learn to align semantically similar instructions despite differences in phrasing. 

For negative samples, we observe that the contrastive loss converges quickly when using instruction-input pairs of different tasks (i.e., normal negatives), leading to minor improvement in robustness. This observation is consistent with the findings in prior studies \citep{liu_how_2023}: LLMs can distinguish between instructions of different tasks such that their hidden representations already have low cosine similarity. To collect \emph{hard negatives}, we draw inspiration from near-OOD samples, which are data that come from the same task but with different classes \cite{winkens_contrastive_2020, fort_exploring_2021, liu_how_2023}. Prior studies found that it is more difficult for models to detect near-OOD samples than samples from other tasks (far-OOD). This finding indicates that the hidden representations of near-OOD samples may not be distinguishable enough and thus can provide informative supervision signals for contrastive learning. 
Accordingly, we choose such a sample $(I_i^-, x_i^-, y_i^-)$ that shares the same instruction as the original instance ($I_i^- = I_i$) but is paired with different input ($x_i^- \neq x_i$) and output ($y_i^- \neq y_i$) as a negative sample. For example, if $y_i$ is "yes", then $y_i^-$ can be "no", ensuring the fundamental intent of the instruction-instance pair is different from the original one. Based on this approach, \MODEL encourages the model to align semantically equivalent instructions with different phrasings while contrasting inputs with different user intents.

\subsection{Learning Objective}
\label{sec/method/contrastive}

Our method \MODEL is illustrated in \Cref{fig:overview}. We construct the training batch such that each original sample is matched with a perturbed instruction and an identical instance as a positive sample. All other in-batch samples are hard negatives selected according to \Cref{section/pos_neg_selection}, i.e. share the same instruction but paired with different instances.

Let $h_i$,  $h_i^+$, and $h_i^-$ indicate model $\mathcal{M}$'s hidden representation of the original, positive, and negative instruction-instance pairs, respectively. Since each original pair may have multiple in-batch negatives, here we use $h_{ij}^-$ to indicate the hidden representation of the negative samples. To align the hidden representation $h_i$ and $h_i^+$, we optimize the model $\mathcal{M}$ with 
the contrastive loss $\mathcal{L}_{ctr}^i$, which is defined as

\begin{equation*}
    \label{eq/contrastive}
    \mathcal{L}_{ctr}^i=
    -\log\frac{e^{\text{sim}(h_i, h_i^+)/\tau}}{e^{\text{sim}(h_i, h_i^+)/\tau} + 
    \sum_j e^{\text{sim}(h_i, h_{ij}^-)/\tau}} ,
\end{equation*}

\noindent
where $\text{sim}(h_1, h_2)$ is the cosine similarity $\frac{h_1^Th_2}{||h_1||\cdot||h_2||}$, and $\tau$ is a temperature hyperparameter. In \MODEL, we obtain the hidden representations by using the hidden state of the last token\footnote{We also experimented with other pooling methods such as max and average pooling but found that using the last token's hidden state yielded better results.} from the decoder of the language model.

To preserve the generation ability of the language model, we follow \citet{liu_brio_2022} to include the standard cross entropy loss for each instruction pair, which can be defined as follows:

\begin{equation*}
    \label{eq/entropy}
    \mathcal{L}^i_{ent} = \frac{1}{l}\sum_{k = 1}^l -\log p(y_k | I_i, x_i, y_{<k})
\end{equation*}

\noindent
where $l$ is the length of the desired output for instruction-input pair $(I_i, x_i)$. This loss is computed for all samples in the batch.

Combining the above two parts, the overall learning objective is
\begin{equation*}
    \label{eq/coin}
    \mathcal{L}^{i}_{\MODEL} = \mathcal{L}^{i}_{ent} + \max(\lambda, \text{detach}(\frac{\mathcal{L}^{i}_{ent}}{\mathcal{L}^{i}_{ctr}})) \mathcal{L}^{i}_{ctr} ,
\end{equation*}
where detach($\cdot$) indicates that the loss value is detached from the computation graph and thus is treated only as a scalar. $\lambda$ is the upper bound of the weight that is assigned to the contrastive loss. Based on empirical results, we found that setting \(\lambda\) too high, thereby significantly increasing the magnitude of the contrastive loss \(\mathcal{L}_{ctr}\) relative to the generation loss \(\mathcal{L}_{ent}\), adversely affects the models' generation ability. To mitigate this issue, we scale the contrastive loss to the same magnitude as the generation loss while setting an upper bound to the weighting, ensuring a balanced influence between enhancing robustness and maintaining generative performance. For more details on the weighting choice of the contrastive loss, refer to \ref{sec/analysis/weighting_of_contrastive_loss}.

\section{Experiment}
In this section, we evaluate \MODEL's performance on enhancing LLMs' robustness to instruction variations on PromptBench, specifically $10$ GLUE datasets with unseen\footnote{In this paper, ``unseen instructions'' refer to those whose textual expressions do not appear in the instruction-tuning dataset. Note that if the model exhibits inadequate robustness when handling unseen instructions for known tasks, its performance is likely to decrease further when confronted with unknown tasks. We consider the former as a rigorous evaluation setting without additional confounding factors.} instructions perturbed at different levels. We first provide an overview of the experiment settings (\Cref{sec/experiment/training_dataset}, \Cref{sec/experiment/implementation}, and \Cref{sec/experiment/evaluation}) and then present a detailed analysis of the experiment results \Cref{sec/experiment/results}. 

\subsection{Training Datasets}
\label{sec/experiment/training_dataset}
In this work, we conduct experiments on a widely used instruction tuning dataset: the FLAN Collection \cite{wei_finetuned_2022}. FLAN Collection \cite{wei_finetuned_2022} is a large-scale data collection that encompasses a wide range of tasks, including natural language inference, common sense reasoning, sentiment analysis, paraphrase identification, etc. This data collection is created by transforming 62 publicly available text datasets into instructional formats. $10$ unique instruction templates are manually composed for each dataset. In this work, we choose 25 datasets with deterministic answers from the collection. To ensure each dataset has an equal chance of being sampled into the training set of \MODEL, we iterate through the training split of each dataset with a round-robin approach. For each entry, we create a positive sample by randomly selecting a predefined instruction template not used by the entry to paraphrase the instruction. Only paraphrasing is used for creating training data while various types of perturbations are included for evaluation (refer to \Cref{sec/experiment/evaluation}). Avoiding assumptions about specific types of noise in instructions is crucial due to the high uncertainty LLMs face in real-world deployment. Hence, a robustness training method that can generalize to other types of perturbations is more desirable. We then select one entry from the remaining dataset as a negative sample, following the strategy in \Cref{section/pos_neg_selection}. Refer to \Cref{section/appendix/dataset} for more details of the processed dataset.

\begin{figure*}
    \centering
    \includegraphics[width=1\linewidth]{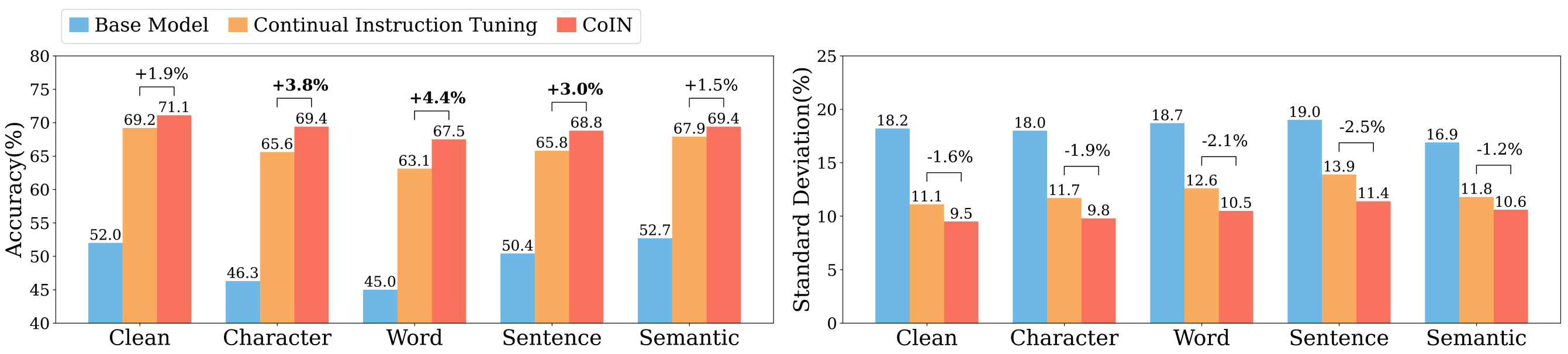}
    \caption{Models' average accuracy (left) and standard deviation (right) across $10$ GLUE datasets, with each dataset having six unseen instructions with no perturbation (clean) or perturbation added at character, word, sentence, and semantic levels. \MODEL has consistent improvement in accuracy and decrease in standard deviation across all types of perturbation compared to the base model and continual instruction tuning. \MODEL obtains significant improvement in robustness against word, character, and sentence level perturbations.}
    \label{fig:main_result}
\end{figure*}

\subsection{Implementation Details}
\label{sec/experiment/implementation}
We use Alpaca \cite{taori_stanford_2023}, a model instruction-tuned from the LLaMA model \cite{touvron_llama_2023} on the 52k Self-Instruct dataset, as the base model. When training models on the augmented FLAN collection, we use the same set of hyper-parameters, with the learning rate, batch size, and cut-off length set to $1*10^{-4}$, 64, and 256 respectively. Since we observe that the magnitude of the contrastive loss can be small during the later phase of training and following \citet{gao_simcse_2022}, we set the temperature $\tau$ and $\lambda$ to $0.05$ and $1000$. All experiments are run on 2 NVIDIA RTX A5000 GPUs. 

\subsection{Evaluation Setting}
\label{sec/experiment/evaluation}
To evaluate models' robustness against variations in expression of instructions, we adopt the PromptBench benchmark \cite{zhu_promptbench_2023}. 
Incorporating a diverse set of tasks, such as sentiment analysis, grammar correctness, duplicate sentence detection, and natural language inference, PromptBench introduces perturbation to task instructions at various levels: character, word, sentence, and semantic. Regarding the data used for evaluation, we sample 300 instruction-instance pairs from each GLUE task wherever the validation set exceeds this size.\footnote{Due to the extensive computational requirement of evaluating the models on the entire benchmark, we sample a subset of instructions and data from all possible instructions and datasets.} For each dataset, PromptBench predefines $20$ instructions. We ensure that all selected and perturbed instructions for each dataset are not seen during the training time. Given that all instructions are unseen while GLUE tasks are seen during training time, this setting allows a more focused evaluation of LLMs' robustness against variations in instructions without the confounding factor of task generalization. 

\stitle{Instruction Variations}
Regarding instructions, we select six clean instructions predefined for each task. Then, we create perturbed versions of each instruction. Following PromptBench, we use DeepWordBug \cite{gao_black-box_2018} to introduce %
character-level perturbations for certain words, and use TextFooler \cite{jin_is_2020} to replace words with contextually similar words. At the sentence level, we implement the CheckList \cite{ribeiro_beyond_2020} and append randomly generated sequences, which all have a length of $10$ and consist of alphabets and digits, at the end of an instruction to distract LLMs. For the semantic-level perturbation, PromptBench defines $10$ instructions that paraphrase the original instruction for each task while following the linguistic behavior of six languages: Chinese, French, Arabic, Spanish, Japanese, and Korean. To keep the number of instructions the same as other types of perturbation, we randomly select one instruction from each language defined for each task, which are all different from the clean instructions. We also ensure that instructions used for evaluation differ from all instructions in the training dataset and thus are unseen by the model, preventing data contamination.

\stitle{Metrics}
For each type of perturbation, we report average accuracy and standard deviations of six instructions created for each GLUE dataset.

\subsection{Results}
\label{sec/experiment/results}

\begin{figure*}[t]
    \centering
    \includegraphics[width=1\linewidth]{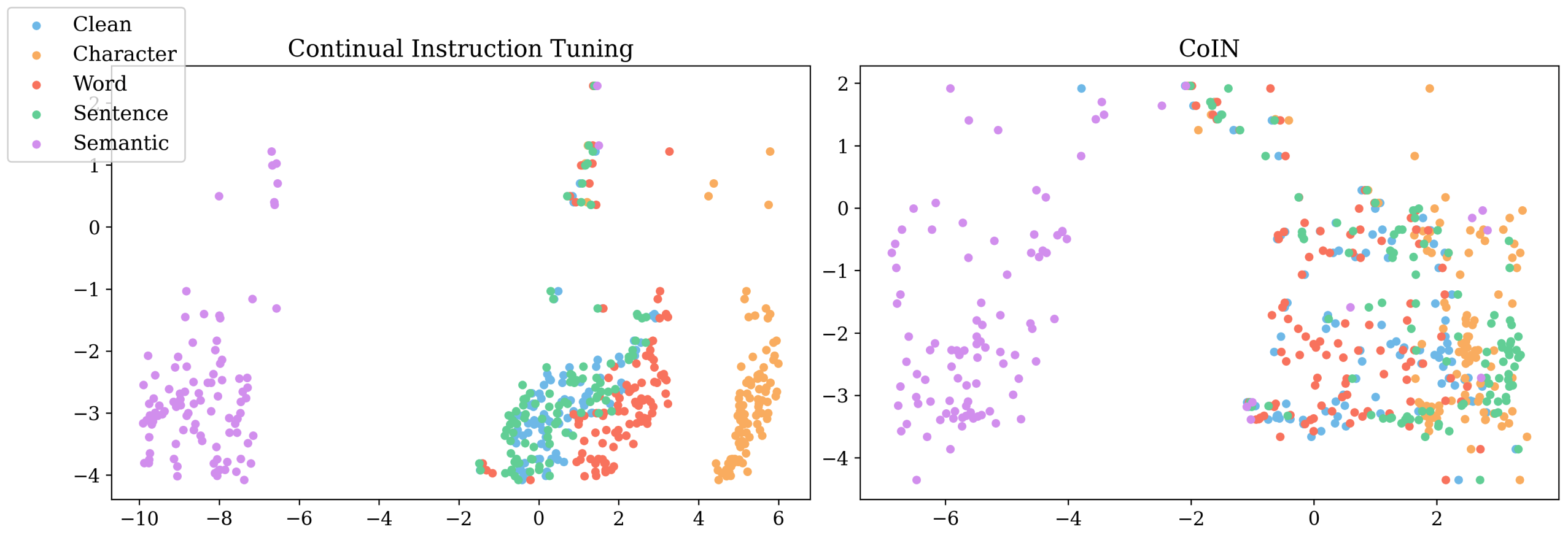}
    \caption{UMAP \cite{mcinnes_umap_2020} visualization of the hidden representations of decoder's last output token from continually instruction-tuned model (left) and \MODEL (right). 300 data points are selected from CoLA \cite{warstadt_neural_2019} with no perturbations (clean) or perturbations added at different levels. \MODEL's representations of inputs with instruction variations are clustered closer to each other compared to the continually instruction-tuned model, especially inputs with perturbations at word, character, and sentence level.
    }
    \label{fig:hidden_states_visualization}
\end{figure*}

In \Cref{fig:main_result}, we evaluate the base model, continual instruction tuning (i.e., base model fine-tuned on the same data as \MODEL with cross entropy loss only), and \MODEL on five groups of instructions across $10$ GLUE datasets. Except for the clean group, which includes the original instructions defined for each dataset, each group contains instructions with the same type of perturbation, including character, word, sentence, and semantic perturbations. 

The base model exhibits low accuracy and large performance variance when given instructions with different perturbations or instructions within the same perturbation group. With only around $52\%$ accuracy on the clean instructions, the base model's performance further decreases when the instructions are perturbed in all character, word, and sentence levels. The largest accuracy gap across different groups is $7.7\%$, observed between the word and the semantic groups. For instructions within the same group, the base model exhibits a variance ranging from $16.9\%$ to $19.0\%$. These observations demonstrate that the base model is sensitive to how instructions are formulated for different tasks. 

Compared to the base model, the continually instruction-tuned model shows increases in accuracy, which is expected as the model is exposed to more data and trained with more steps. Nevertheless, the performance gap between different groups can still be as large as $6.1\%$ observed between the clean group and the group with word-level perturbation. This shows that the continually instruction-tuned model still lacks robustness to unseen instructions with variations across different levels. 

Compared to continual instruction tuning, \MODEL further reduces performance variance and consistently improves accuracy for instructions within and across different groups without introducing any new data and training steps. As it can be seen from \Cref{fig:main_result}, \MODEL achieves improvements in accuracy for all types of perturbation, up to $4.4\%$ for word-level perturbations where the continually instruction-tuned model exhibits its largest drop in performance. %
The largest performance gap is reduced to $3.6\%$. The consistent improvement across all types of perturbations demonstrates the generalizability of \MODEL at enhancing models' robustness against variations in instructions at different levels. \MODEL also decreases the performance variance on instructions from the five groups by $1.6\%$, $1.9\%$, $2.1\%$, $2.5\%$, and $1.2\%$. This also shows that \MODEL can effectively help the model become less sensitive to specific instructions for each task and more consistent in its performance. For more detailed results, refer to \Cref{tab:on_flan_detail_accuracy}.

\section{Analyses}
\label{sec/analysis}

To provide a more comprehensive view of the impact of \MODEL on the model's robustness to instruction variations, we further analyze the results of our method by examining the hidden representations of instruction variants (\Cref{sec/analysis/closer_representations_of_instruction_variants}), task category (\Cref{sec/analysis/impact_on_different_tasks}), and the weighting choice for the contrastive loss (\Cref{sec/analysis/weighting_of_contrastive_loss}).

\subsection{Closer Representations of Instruction Variants}
\label{sec/analysis/closer_representations_of_instruction_variants}

To understand the impact of \MODEL on the representations of instructions with variations at different levels, we visualize the hidden states of the last output tokens from the decoder's last Transformer layer. Specifically, we select 300 data points from CoLA \cite{warstadt_neural_2019}, choose one of its instructions, add perturbations at different levels to the instruction, and obtain the hidden states from the model. 

As observed in \Cref{fig:hidden_states_visualization}, \MODEL's hidden representations of inputs with instruction variations at different levels are much closer than those of the continually instruction-tuned model. In the embedding space of the continually instruction-tuned model, the representation of instructions with different perturbations, especially at character and word levels, are clustered almost into distinct groups. This may indicate that the model treats data points with the same instruction varied at different levels differently and thus is more sensitive to how the same instruction is formulated. 

In contrast, the representations of data points with character, word, and sentence level variations are less distinguishable in \MODEL's embedding space, with representations of instructions varied at the word level (red) having greater overlap with those of the clean group (blue). This observation can be associated with \MODEL's varied improvement in performance across different perturbations. As shown in \Cref{fig:main_result}, \MODEL achieves more evident improvement on instructions with word, character, and sentence level perturbations. It can be concluded from the two figures that when \MODEL effectively aligns the representations of perturbed instructions to those of the clean instructions, the model becomes more capable of capturing the original semantic meaning of the instructions. Thus, it becomes less sensitive to perturbations in instructions. 

It can be observed that the representations of instructions with semantic level perturbation are located relatively far away from those of instructions with other types of perturbation. This is expected as paraphrasing introduces new structure and wording to the original instruction, which may lead to varied hidden representations. Nonetheless, \MODEL %
stabilizes the representation of the original and paraphrased instructions, demonstrating \MODEL can effectively align the representation of instruction variants with each other and thus enhance the model's robustness to instruction variations. 

\subsection{Impact on Different Tasks}
\label{sec/analysis/impact_on_different_tasks}

\begin{table*}
\centering
\begin{tabular}{c|cc|cc|cc}
\toprule
(\%)& \multicolumn{2}{c}{Continual Instruction Tuning } & \multicolumn{2}{c}{\MODEL} & \multicolumn{2}{c}{$\triangle$} \\
\midrule
Task & Accuracy & Std & Accuracy & Std & Accuracy & Std \\
\midrule
Sentiment Analysis & 89.0 & 4.1 & 90.4 & 3.1 & \textbf{+1.4} & \textbf{-1.1} \\
Natural Language Inference & 64.4 & 3.7 & 66.1 & 3.5 & \textbf{+1.7} & -0.2 \\
Paraphrase Identification & 63.0 & 11.0 & 68.5 & 5.9 & \textbf{+5.4} & \textbf{-5.1} \\
Grammar Correctness & 62.0 & 9.2 & 68.4 & 3.9 & \textbf{+6.3} & \textbf{-5.3} \\
\bottomrule
\end{tabular}
\label{tab:change_in_accuracy_ablation_coin}
\caption{Models' average accuracy and standard deviation of each task category. \MODEL has consistent improvement across all tasks with more evident improvement on duplicate sentence detection and grammar correctness tasks.}
\end{table*}

We examine \MODEL's influence on the model's performance for different tasks. Based on the task category defined in the PromptBench benchmark, we split the $10$ GLUE datasets into four categories: (1) sentiment analysis, (2) natural language inference (NLI), (3) paraphrase identification, and (4) grammar correctness. Refer to \Cref{tab:glue_task_category} for specific datasets classified to each category. 

As shown in \Cref{tab:change_in_accuracy_ablation_coin}, \MODEL achieves evident improvements in accuracy by +5.4\% and +6.3\% on paraphrase identification and grammar correctness tasks. Intuitively, these tasks can benefit more directly from \MODEL that aims to enhance the similarity of representations of semantically equivalent instruction-input pairs. For example, paraphrase identification can directly benefit from the model's more refined ability to group textual inputs with similar semantic meanings, as \MODEL pushes representations of inputs with different meanings away from each other. Similarly, grammar correctness can also benefit from the contrastive loss, which may group hidden representations of grammatically correct inputs closer to each other and thus enable the model to become better at detecting inputs with invalid syntactic structures and grammatical rules. 

On the other hand, \MODEL gains modest enhancement in accuracy on sentiment analysis and NLI tasks by +1.4\% and +1.7\% compared to the continual instruction-tuned model. For the sentiment analysis task, the continually instruction-tuned model has already achieved an accuracy of 89.0\%. Obtaining further improvements can be challenging given that the model is already capable at distinguishing between sentences with different sentiments. Regarding NLI, the task requires a comprehensive understanding of the relationship between two sentences, which can depend on the model's knowledge of various domains or reasoning ability to infer implicit meanings that are not directly stated. The complex relation between two sentences may not be explicitly captured by the hidden representations, meaning that \MODEL may not explicitly further enhance the model's reasoning ability. However, \MODEL still obtains an improvement of +1.4\% and +1.7\% on the two tasks, demonstrating \MODEL's effectiveness at enhancing the model's ability to discern the nuanced inferential relation that underlies the overall semantic meaning of the instruction-input pairs.

\begin{table*}
    \scriptsize
    \setlength{\tabcolsep}{1pt}
    \centering
    \begin{tabular}{c|c|c|c|c|c|c|c|c|c|c|c|c}
    \toprule
    Model & Perturbation & CoLA & MNLI & MNLI-m & MNLI-mm & MRPC & QNLI & QQP & RTE & SST2 & WNLI & Average \\
    \midrule
    \multirow{5}{*}{ \makecell{Alpaca \\ Baseline} } & Clean & 65.1 $\pm$ 2.1 & 51.5 $\pm$ 4.3 & 51.5 $\pm$ 4.3 & 51.3 $\pm$ 5.0 & 28.6 $\pm$ 27.5 & 51.8 $\pm$ 1.5 & 26.6 $\pm$ 10.8 & 62.2 $\pm$ 2.4 & 80.9 $\pm$ 5.7 & 50.5 $\pm$ 3.4 & 52.0 $\pm$ 18.2 \\
    & Character & 61.8 $\pm$ 4.6 & 47.2 $\pm$ 6.4 & 47.2 $\pm$ 6.4 & 49.3 $\pm$ 4.5 & 27.4 $\pm$ 24.1 & 42.7 $\pm$ 6.9 & 15.6 $\pm$ 10.9 & 55.5 $\pm$ 5.6 & 66.7 $\pm$ 15.6 & 49.3 $\pm$ 3.5 & 46.3 $\pm$ 18.0 \\
    & Word & 61.7 $\pm$ 2.0 & 49.6 $\pm$ 3.8 & 49.6 $\pm$ 3.8 & 49.2 $\pm$ 4.7 & 43.3 $\pm$ 21.8 & 24.8 $\pm$ 17.4 & 14.7 $\pm$ 8.1 & 57.5 $\pm$ 4.9 & 46.4 $\pm$ 25.8 & 53.1 $\pm$ 2.7 & 45.0 $\pm$ 18.7 \\
    & Sentence & 64.8 $\pm$ 1.8 & 51.2 $\pm$ 3.6 & 51.2 $\pm$ 3.6 & 52.9 $\pm$ 2.2 & 15.3 $\pm$ 10.7 & 50.2 $\pm$ 3.2 & 22.6 $\pm$ 6.8 & 61.5 $\pm$ 3.3 & 82.3 $\pm$ 4.1 & 52.1 $\pm$ 2.0 & 50.4 $\pm$ 19.0 \\
    & Semantic & 65.4 $\pm$ 1.9 & 52.1 $\pm$ 1.2 & 52.1 $\pm$ 1.2 & 51.6 $\pm$ 1.8 & 37.9 $\pm$ 25.6 & 52.1 $\pm$ 3.7 & 25.8 $\pm$ 10.0 & 59.2 $\pm$ 4.4 & 82.1 $\pm$ 3.3 & 48.6 $\pm$ 4.4 & 52.7 $\pm$ 16.9 \\
    \midrule
    \multirow{5}{*}{ \makecell{Continual \\Instruction \\ Tuning} } & Clean & 63.5 $\pm$ 8.6 & 68.7 $\pm$ 2.4 & 67.3 $\pm$ 2.7 & 66.3 $\pm$ 2.7 & 62.8 $\pm$ 13.0 & 62.9 $\pm$ 4.2 & 71.2 $\pm$ 7.6 & 82.0 $\pm$ 1.9 & 90.1 $\pm$ 2.4 & 57.5 $\pm$ 3.8 & 69.2 $\pm$ 11.1 \\
    & Character & 64.9 $\pm$ 3.1 & 64.9 $\pm$ 2.1 & 64.1 $\pm$ 2.3 & 63.4 $\pm$ 1.9 & 62.1 $\pm$ 11.9 & 54.7 $\pm$ 3.6 & 61.9 $\pm$ 11.8 & 75.7 $\pm$ 4.8 & 90.5 $\pm$ 2.0 & 54.0 $\pm$ 5.1 & 65.6 $\pm$ 11.7 \\
    & Word & 58.9 $\pm$ 12.6 & 64.8 $\pm$ 4.1 & 65.4 $\pm$ 3.8 & 64.3 $\pm$ 3.5 & 56.4 $\pm$ 10.5 & 46.8 $\pm$ 6.7 & 62.5 $\pm$ 8.2 & 73.8 $\pm$ 3.5 & 84.2 $\pm$ 12.6 & 54.0 $\pm$ 2.1 & 63.1 $\pm$ 12.6 \\
    & Sentence & 58.6 $\pm$ 15.2 & 66.4 $\pm$ 1.9 & 65.3 $\pm$ 1.4 & 65.1 $\pm$ 3.7 & 55.9 $\pm$ 16.8 & 53.2 $\pm$ 8.6 & 66.6 $\pm$ 8.1 & 80.3 $\pm$ 3.0 & 90.4 $\pm$ 1.2 & 55.9 $\pm$ 4.3 & 65.8 $\pm$ 13.9 \\
    & Semantic & 64.3 $\pm$ 6.6 & 67.0 $\pm$ 2.9 & 67.1 $\pm$ 2.5 & 66.0 $\pm$ 3.1 & 61.4 $\pm$ 14.3 & 56.4 $\pm$ 9.9 & 69.6 $\pm$ 8.1 & 80.0 $\pm$ 4.4 & 89.6 $\pm$ 2.5 & 58.0 $\pm$ 4.6 & 67.9 $\pm$ 11.8 \\
    \midrule
    \multirow{5}{*}{ \MODEL } & Clean & 70.4 $\pm$ 3.9 & 68.8 $\pm$ 2.7 & 68.0 $\pm$ 2.2 & 67.6 $\pm$ 3.5 & 70.6 $\pm$ 3.5 & 61.9 $\pm$ 6.0 & 70.1 $\pm$ 6.0 & 82.3 $\pm$ 1.5 & 91.4 $\pm$ 0.7 & 59.9 $\pm$ 2.5 & 71.1 $\pm$ 9.5 \\
    & Character & 66.9 $\pm$ 3.0 & 68.2 $\pm$ 2.0 & 67.5 $\pm$ 1.3 & 66.6 $\pm$ 4.0 & 72.4 $\pm$ 2.5 & 58.7 $\pm$ 4.2 & 64.7 $\pm$ 8.0 & 78.5 $\pm$ 3.1 & 91.1 $\pm$ 2.1 & 58.9 $\pm$ 2.6 & 69.4 $\pm$ 9.8 \\
    & Word & 66.5 $\pm$ 4.5 & 67.4 $\pm$ 1.7 & 67.7 $\pm$ 3.0 & 66.1 $\pm$ 2.3 & 71.9 $\pm$ 5.4 & 49.9 $\pm$ 7.5 & 63.9 $\pm$ 6.0 & 75.6 $\pm$ 3.5 & 85.6 $\pm$ 11.6 & 60.1 $\pm$ 3.8 & 67.5 $\pm$ 10.5 \\
    & Sentence & 68.4 $\pm$ 7.2 & 67.7 $\pm$ 3.5 & 68.2 $\pm$ 2.6 & 66.3 $\pm$ 3.6 & 63.3 $\pm$ 9.6 & 55.4 $\pm$ 9.5 & 66.8 $\pm$ 6.1 & 79.8 $\pm$ 3.5 & 92.3 $\pm$ 0.6 & 59.6 $\pm$ 2.8 & 68.8 $\pm$ 11.4 \\
    & Semantic & 69.7 $\pm$ 1.2 & 66.3 $\pm$ 1.8 & 67.0 $\pm$ 0.5 & 64.3 $\pm$ 2.6 & 72.6 $\pm$ 5.8 & 56.1 $\pm$ 10.0 & 68.5 $\pm$ 6.3 & 78.5 $\pm$ 4.5 & 91.6 $\pm$ 0.6 & 59.2 $\pm$ 2.0 & 69.4 $\pm$ 10.6 \\
    \bottomrule
    \end{tabular}
    \caption{Model's average accuracy and standard deviation on 10 GLUE datasets, each having six instructions with different types of perturbation. \MODEL here is trained with $\lambda = 1,000$.}
    \label{tab:on_flan_detail_accuracy}
\end{table*}

\subsection{Weighting of Contrastive Loss}
\label{sec/analysis/weighting_of_contrastive_loss}
\begin{figure}
    \centering
    \includegraphics[width=1\linewidth]{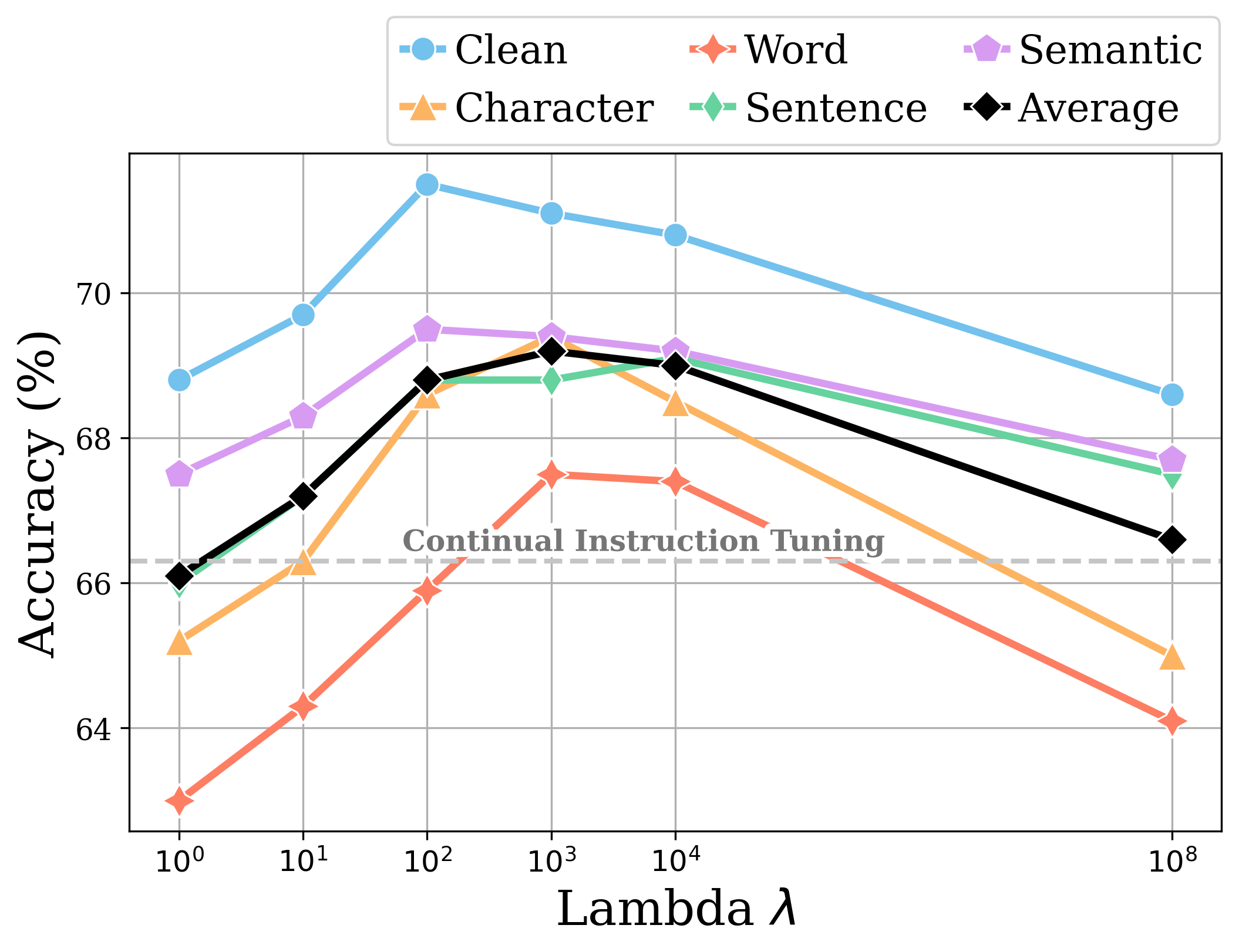}
    \caption{\MODEL's performance by the maximum weight $\lambda$ assigned to the contrastive loss. \MODEL achieves the highest average accuracy at $\lambda=10^3$.}
    \label{fig:contra_loss_weighting_choice}
\end{figure}

As the weight of the contrastive loss may affect the extent to which \MODEL align representations of instruction variants \cite{liu_brio_2022}, we examine how different values assigned to $\lambda$ can affect \MODEL's performance across different perturbation levels. 

As shown in \Cref{fig:contra_loss_weighting_choice}, \MODEL achieves its best average performance when $\lambda=1,000$. When $\lambda$ is small, contrastive loss does not have significant impact on the model due to its small magnitude. The resulting model has similar performance and sensitivity to instruction variations as the continual instruction-tuned model. As $\lambda$ increases, \MODEL's performance increases across different types of perturbations, indicating that the contrastive loss is guiding the model to align representations of instruction variations closer to each other and thus become more robust to the introduced perturbations. 

However, when $\lambda$ is too large, \MODEL's performance decreases significantly, %
Therefore, based on the empirical results, we choose $\lambda=1,000$ for higher accuracy and smaller standard deviation. Refer to \Cref{tab:lambda_detailed_accuracy} for detailed experiment results of models with different contrastive loss weighting.

\section{Conclusion}
In this paper, we proposed \MODEL that aligns hidden representations of semantically equivalent instruction-instance pairs. Evaluation results on PromptBench, with instructions that differ at character, word, sentence, and semantic levels, demonstrate \MODEL's effectiveness of enhancing LLMs' robustness to semantic-invariant instruction variations. Future work can apply contrastive instruction tuning to enhance the robustness of models and tasks in other modalities, and on other prompt components such as few-shot demonstrations and system prompts.%

\section*{Limitation}
We summarize the limitations of this work as follows:
First, the current contrastive data selection method only considers paraphrasing for positive instruction augmentation. More semantic-invariant data augmentation methods could be explored.
Second, the experiment scale could be enlarged to include more instruction tuning datasets, instruction-tuned models, and downstream tasks. This would provide additional evidence about \MODEL's effectiveness.
Third, while we use a rigorous evaluation setting to measure model robustness, evaluating the influence of \MODEL from other perspectives could enhance understanding of contrastive instruction tuning.

\section*{Acknowledgement}

We appreciate the reviewers for their insightful
comments and suggestions.
Tianyi Yan was supported by the Center for Undergraduate Research in Viterbi Engineering (CURVE) Fellowship.
Fei Wang was supported by the Amazon ML Fellowship.
James Y. Huang was supported by a gift fund from the Amazon Center on Secure \& Trusted ML.
Muhao Chen was supported by the NSF Grant IIS 2105329, the NSF Grant ITE 2333736, the DARPA AIE Grant HR0011-24-9-0370, and an Amazon Research Award.

\bibliography{reference}
\bibliographystyle{acl_natbib}
\clearpage

\appendix

\section{Datasets}
\label{section/appendix/dataset}

\begin{table*}
    \centering
    \begin{tabular}{c|c|c}
    \toprule
    Task Category & Dataset & Count \\
    \midrule
    \multirow{10}{*}{ Natural Language Inference(NLI) } & ANLI(R1) & 2664 \\
    & ANLI(R2) & 2670 \\
    & ANLI(R3) & 2658 \\
    & CB & 232 \\
    & MNLI-Matched & 2678 \\
    & MNLI-Mismatched & 2678 \\
    & QNLI & 2682 \\
    & RTE & 2328 \\
    & SNLI & 2682 \\
    & WNLI & 920 \\
    \midrule
    \multirow{4}{*}{ Sentiment Analysis } & IMDB & 354 \\
    & Sent140 & 2684 \\
    & SST2 & 2682 \\
    & Yelp & 834 \\
    \midrule
    \multirow{4}{*}{ Paraphrase Identification} & MRPC & 2684 \\
    & QQP & 2684 \\
    & PAWS Wiki & 2684 \\
    & STS-B & 2682 \\
    \midrule
    \multirow{2}{*}{ Reading Comprehension } & BoolQ & 1044 \\
    & MultiRC & 30 \\
    \midrule
    Coreference Resolution & WSC273 & 720 \\
    \midrule
    Summarization & AG News & 2678 \\
    \midrule
    \multirow{3}{*}{ Miscellaneous } & TREC & 2682 \\
    & CoLA & 2684 \\
    & WIC & 2684 \\
    \midrule
    Total & & 52002 \\
    \bottomrule
    \end{tabular}
    \caption{Number of entries sampled for each dataset from the FLAN collection}
    \label{tab:flan_dataset_cnt}
\end{table*}

For the training dataset sampled from the FLAN collection released under Apache-2.0 license, we select 25 datasets with answer options, which can be classified into 7 categories:
\begin{enumerate}
    \item Natural Language Inference (NLI): how two sentences are related. The following datasets are used:
    \begin{enumerate}
        \item ANLI \cite{nie_adversarial_2020} 
        \item CB \cite{marneffe_commitmentbank_2019}
        \item MNLI \cite{williams_broad-coverage_2018}
        \item QNLI \cite{rajpurkar_know_2018}
        \item RTE \citep{dagan_pascal_2006, bar-haim_second_2006, giampiccolo_third_2007, bentivogli_fifth_2009}
    \end{enumerate}
    \item Sentiment Analysis: whether the input text has positive or negative sentiment. The following datasets are used:
    \begin{enumerate}
        \item IMDB \cite{maas_learning_2011}
        \item Sent140 \cite{go_twitter_2009}
        \item SST2 \cite{socher_recursive_2013}
        \item Yelp \cite{zhang_character-level_2015}
    \end{enumerate}
    \item Paraphrase Detection: whether two sentences are semantically equivalent. The following datasets are used:
    \begin{enumerate}
        \item MRPC \cite{dolan_automatically_2005}
        \item QQP \cite{wang_glue_2018}
        \item Paws Wiki \cite{zhang_paws_2019}
        \item STS-B \cite{cer_semeval-2017_2017}
    \end{enumerate}
    \item Reading Comprehension: answer questions based on passages that contain the answers. The following datasets are used:
    \begin{enumerate}
        \item BoolQ \cite{clark_boolq_2019}
        \item MultiRC \cite{khashabi_looking_2018}
    \end{enumerate}
    \item Coreference: find expressions that refer to the same entity in the input text. WSC273 dataset is used \cite{levesque_winograd_2012}. 
    \item Summarization: produce an abbreviated summary of the input text. For input with answer options, the model is asked to, for instance, choose the broader topic or the best summary among all choices provided. AG news dataaset is used \cite{zhang_character-level_2015}.
    \item Miscellaneous:
        \begin{enumerate}
            \item TREC \citep{li_learning_2002, hovy_toward_2001}: Classify questions into specified categories, such as whether the question is related to human, location, abbreviations, etc.
            \item CoLA \cite{warstadt_neural_2019}: Linguistic acceptability.
            \item WIC \cite{pilehvar_wic_2019}: Evaluate intended meaning of a word within a context.
        \end{enumerate}
\end{enumerate}

Refer to \Cref{tab:flan_dataset_cnt} for number of entries filtered and selected out from each dataset following the rules described in \cref{sec/experiment/training_dataset}.

\section{Detailed Experiment Results}
\begin{table*}[t]
    \scriptsize
    \setlength{\tabcolsep}{1pt}
    \centering
    \begin{tabular}{c|c|c|c|c|c|c|c|c|c|c|c|c}
    \toprule
    $\text{Lambda } \lambda$ & Perturbation & CoLA & MNLI & MNLI-m & MNLI-mm & MRPC & QNLI & QQP & RTE & SST2 & WNLI & Average \\
    \midrule
    \multirow{5}{*}{ 1 } & Clean & 66.4 $\pm$ 6.0 & 67.7 $\pm$ 2.6 & 67.8 $\pm$ 2.6 & 65.8 $\pm$ 1.4 & 63.6 $\pm$ 15.2 & 62.3 $\pm$ 5.5 & 66.4 $\pm$ 12.1 & 81.7 $\pm$ 2.9 & 90.1 $\pm$ 1.7 & 56.6 $\pm$ 3.9 & 68.8 $\pm$ 11.6 \\
    & DeepWordBug & 65.0 $\pm$ 3.4 & 65.2 $\pm$ 1.7 & 64.6 $\pm$ 1.8 & 63.3 $\pm$ 2.0 & 63.3 $\pm$ 11.3 & 54.7 $\pm$ 3.5 & 57.6 $\pm$ 11.2 & 75.3 $\pm$ 3.6 & 90.3 $\pm$ 1.8 & 52.8 $\pm$ 4.2 & 65.2 $\pm$ 11.8 \\
    & TextFooler & 58.7 $\pm$ 11.0 & 65.4 $\pm$ 1.9 & 66.2 $\pm$ 2.8 & 64.3 $\pm$ 3.7 & 59.9 $\pm$ 10.7 & 46.8 $\pm$ 6.0 & 55.6 $\pm$ 12.0 & 74.1 $\pm$ 4.1 & 85.0 $\pm$ 13.4 & 54.0 $\pm$ 3.1 & 63.0 $\pm$ 13.0 \\
    & CheckList & 61.3 $\pm$ 13.0 & 67.7 $\pm$ 1.8 & 66.8 $\pm$ 1.9 & 64.3 $\pm$ 3.2 & 57.8 $\pm$ 17.3 & 51.3 $\pm$ 10.0 & 61.9 $\pm$ 12.6 & 80.5 $\pm$ 2.4 & 91.1 $\pm$ 1.6 & 57.5 $\pm$ 2.8 & 66.0 $\pm$ 14.2 \\
    & Semantic & 68.8 $\pm$ 3.6 & 65.1 $\pm$ 1.7 & 65.4 $\pm$ 1.6 & 64.9 $\pm$ 3.2 & 62.6 $\pm$ 15.8 & 56.5 $\pm$ 7.8 & 65.8 $\pm$ 10.3 & 79.6 $\pm$ 2.4 & 89.9 $\pm$ 1.9 & 56.3 $\pm$ 5.2 & 67.5 $\pm$ 11.9 \\
    \midrule
    \multirow{5}{*}{ 10 } & Clean & 69.6 $\pm$ 3.2 & 65.8 $\pm$ 2.1 & 65.4 $\pm$ 2.7 & 64.6 $\pm$ 2.2 & 71.7 $\pm$ 8.1 & 62.5 $\pm$ 5.2 & 68.7 $\pm$ 9.5 & 81.7 $\pm$ 2.9 & 90.0 $\pm$ 2.4 & 56.8 $\pm$ 3.0 & 69.7 $\pm$ 10.4 \\
    & DeepWordBug & 66.3 $\pm$ 2.5 & 64.8 $\pm$ 1.8 & 64.9 $\pm$ 1.5 & 61.3 $\pm$ 1.6 & 70.4 $\pm$ 6.6 & 55.4 $\pm$ 4.0 & 57.4 $\pm$ 7.2 & 76.5 $\pm$ 3.6 & 89.4 $\pm$ 2.6 & 56.8 $\pm$ 4.7 & 66.3 $\pm$ 10.7 \\
    & TextFooler & 61.2 $\pm$ 9.6 & 63.5 $\pm$ 1.8 & 64.6 $\pm$ 1.6 & 62.8 $\pm$ 3.6 & 70.2 $\pm$ 8.2 & 48.4 $\pm$ 5.1 & 56.0 $\pm$ 11.2 & 74.4 $\pm$ 3.9 & 84.2 $\pm$ 12.9 & 57.3 $\pm$ 1.9 & 64.3 $\pm$ 12.0 \\
    & CheckList & 67.6 $\pm$ 8.0 & 66.1 $\pm$ 1.7 & 66.9 $\pm$ 2.2 & 62.6 $\pm$ 2.0 & 64.8 $\pm$ 17.0 & 53.2 $\pm$ 10.4 & 61.4 $\pm$ 11.1 & 80.3 $\pm$ 2.6 & 90.9 $\pm$ 2.2 & 58.2 $\pm$ 2.7 & 67.2 $\pm$ 13.0 \\
    & Semantic & 69.4 $\pm$ 1.3 & 63.7 $\pm$ 1.5 & 64.4 $\pm$ 1.3 & 63.1 $\pm$ 2.6 & 69.7 $\pm$ 10.3 & 57.2 $\pm$ 7.1 & 67.4 $\pm$ 8.7 & 79.5 $\pm$ 2.7 & 89.8 $\pm$ 2.4 & 58.5 $\pm$ 4.1 & 68.3 $\pm$ 10.7 \\
    \midrule
    \multirow{5}{*}{ 100 } & Clean & 69.3 $\pm$ 3.2 & 68.9 $\pm$ 1.7 & 69.1 $\pm$ 1.9 & 66.8 $\pm$ 3.1 & 73.6 $\pm$ 3.8 & 62.3 $\pm$ 5.9 & 70.1 $\pm$ 7.8 & 82.4 $\pm$ 1.6 & 90.6 $\pm$ 1.1 & 62.0 $\pm$ 3.2 & 71.5 $\pm$ 9.2 \\
    & DeepWordBug & 66.5 $\pm$ 3.8 & 68.4 $\pm$ 1.8 & 68.7 $\pm$ 1.6 & 65.5 $\pm$ 2.9 & 73.5 $\pm$ 2.7 & 55.2 $\pm$ 4.3 & 61.9 $\pm$ 8.4 & 77.3 $\pm$ 3.6 & 91.1 $\pm$ 2.1 & 57.5 $\pm$ 2.5 & 68.6 $\pm$ 10.6 \\
    & TextFooler & 62.1 $\pm$ 6.6 & 66.8 $\pm$ 2.9 & 67.5 $\pm$ 2.3 & 66.0 $\pm$ 1.5 & 72.1 $\pm$ 4.9 & 48.5 $\pm$ 7.4 & 60.3 $\pm$ 9.6 & 73.7 $\pm$ 4.5 & 85.8 $\pm$ 10.9 & 56.3 $\pm$ 2.8 & 65.9 $\pm$ 11.5 \\
    & CheckList & 68.9 $\pm$ 5.4 & 69.2 $\pm$ 3.0 & 69.4 $\pm$ 2.8 & 66.3 $\pm$ 3.7 & 64.9 $\pm$ 12.8 & 53.8 $\pm$ 10.0 & 66.1 $\pm$ 8.8 & 80.6 $\pm$ 3.1 & 91.6 $\pm$ 0.7 & 57.0 $\pm$ 2.4 & 68.8 $\pm$ 12.1 \\
    & Semantic & 68.7 $\pm$ 2.1 & 66.9 $\pm$ 1.7 & 67.0 $\pm$ 2.5 & 64.0 $\pm$ 2.4 & 72.3 $\pm$ 6.8 & 55.0 $\pm$ 9.6 & 70.7 $\pm$ 6.7 & 79.8 $\pm$ 3.5 & 91.1 $\pm$ 0.7 & 59.2 $\pm$ 4.7 & 69.5 $\pm$ 10.9 \\
    \midrule
    \multirow{5}{*}{ 1000 } & Clean & 70.4 $\pm$ 3.9 & 68.8 $\pm$ 2.7 & 68.0 $\pm$ 2.2 & 67.6 $\pm$ 3.5 & 70.6 $\pm$ 3.5 & 61.9 $\pm$ 6.0 & 70.1 $\pm$ 6.0 & 82.3 $\pm$ 1.5 & 91.4 $\pm$ 0.7 & 59.9 $\pm$ 2.5 & 71.1 $\pm$ 9.5 \\
    & DeepWordBug & 66.9 $\pm$ 3.0 & 68.2 $\pm$ 2.0 & 67.5 $\pm$ 1.3 & 66.6 $\pm$ 4.0 & 72.4 $\pm$ 2.5 & 58.7 $\pm$ 4.2 & 64.7 $\pm$ 8.0 & 78.5 $\pm$ 3.1 & 91.1 $\pm$ 2.1 & 58.9 $\pm$ 2.6 & 69.4 $\pm$ 9.8 \\
    & TextFooler & 66.5 $\pm$ 4.5 & 67.4 $\pm$ 1.7 & 67.7 $\pm$ 3.0 & 66.1 $\pm$ 2.3 & 71.9 $\pm$ 5.4 & 49.9 $\pm$ 7.5 & 63.9 $\pm$ 6.0 & 75.6 $\pm$ 3.5 & 85.6 $\pm$ 11.6 & 60.1 $\pm$ 3.8 & 67.5 $\pm$ 10.5 \\
    & CheckList & 68.4 $\pm$ 7.2 & 67.7 $\pm$ 3.5 & 68.2 $\pm$ 2.6 & 66.3 $\pm$ 3.6 & 63.3 $\pm$ 9.6 & 55.4 $\pm$ 9.5 & 66.8 $\pm$ 6.1 & 79.8 $\pm$ 3.5 & 92.3 $\pm$ 0.6 & 59.6 $\pm$ 2.8 & 68.8 $\pm$ 11.4 \\
    & Semantic & 69.7 $\pm$ 1.2 & 66.3 $\pm$ 1.8 & 67.0 $\pm$ 0.5 & 64.3 $\pm$ 2.6 & 72.6 $\pm$ 5.8 & 56.1 $\pm$ 10.0 & 68.5 $\pm$ 6.3 & 78.5 $\pm$ 4.5 & 91.6 $\pm$ 0.6 & 59.2 $\pm$ 2.0 & 69.4 $\pm$ 10.6 \\
    \midrule
    \multirow{5}{*}{ 10000 } & Clean & 69.6 $\pm$ 5.5 & 67.9 $\pm$ 2.4 & 68.6 $\pm$ 2.1 & 67.4 $\pm$ 1.7 & 69.0 $\pm$ 8.5 & 63.9 $\pm$ 6.0 & 72.9 $\pm$ 5.9 & 81.1 $\pm$ 2.2 & 91.3 $\pm$ 0.9 & 56.8 $\pm$ 4.7 & 70.8 $\pm$ 10.1 \\
    & DeepWordBug & 66.4 $\pm$ 3.7 & 67.2 $\pm$ 2.7 & 67.4 $\pm$ 2.0 & 66.9 $\pm$ 3.5 & 64.3 $\pm$ 8.0 & 59.8 $\pm$ 4.4 & 65.9 $\pm$ 9.0 & 77.2 $\pm$ 2.4 & 90.7 $\pm$ 2.7 & 58.5 $\pm$ 2.7 & 68.5 $\pm$ 10.0 \\
    & TextFooler & 62.9 $\pm$ 7.9 & 66.7 $\pm$ 2.7 & 66.5 $\pm$ 2.7 & 65.6 $\pm$ 2.7 & 68.4 $\pm$ 9.4 & 54.8 $\pm$ 7.3 & 66.8 $\pm$ 6.3 & 76.2 $\pm$ 3.6 & 84.8 $\pm$ 11.5 & 61.0 $\pm$ 3.5 & 67.4 $\pm$ 10.1 \\
    & CheckList & 68.9 $\pm$ 7.9 & 67.2 $\pm$ 2.9 & 67.4 $\pm$ 2.8 & 65.4 $\pm$ 2.4 & 61.7 $\pm$ 17.6 & 59.2 $\pm$ 9.0 & 70.5 $\pm$ 6.6 & 79.7 $\pm$ 3.1 & 92.2 $\pm$ 0.5 & 58.7 $\pm$ 3.8 & 69.1 $\pm$ 12.1 \\
    & Semantic & 69.5 $\pm$ 2.8 & 65.9 $\pm$ 2.1 & 66.1 $\pm$ 2.3 & 65.5 $\pm$ 2.2 & 67.2 $\pm$ 13.4 & 60.1 $\pm$ 7.7 & 70.7 $\pm$ 6.6 & 77.9 $\pm$ 4.6 & 91.4 $\pm$ 0.9 & 58.0 $\pm$ 1.5 & 69.2 $\pm$ 10.7 \\
    \midrule
    \multirow{5}{*}{ 100000000 } & Clean & 70.4 $\pm$ 3.0 & 66.2 $\pm$ 2.1 & 66.1 $\pm$ 1.9 & 65.7 $\pm$ 1.7 & 55.0 $\pm$ 10.3 & 61.2 $\pm$ 7.3 & 70.9 $\pm$ 4.9 & 83.3 $\pm$ 1.1 & 90.6 $\pm$ 1.5 & 56.6 $\pm$ 3.7 & 68.6 $\pm$ 11.5 \\
    & DeepWordBug & 64.4 $\pm$ 4.5 & 63.4 $\pm$ 3.0 & 63.2 $\pm$ 2.7 & 64.1 $\pm$ 2.0 & 46.2 $\pm$ 4.3 & 60.3 $\pm$ 5.8 & 64.6 $\pm$ 6.0 & 80.3 $\pm$ 2.4 & 86.7 $\pm$ 6.3 & 56.3 $\pm$ 4.0 & 65.0 $\pm$ 11.6 \\
    & TextFooler & 62.9 $\pm$ 8.1 & 64.3 $\pm$ 3.7 & 62.7 $\pm$ 3.6 & 63.9 $\pm$ 3.4 & 49.1 $\pm$ 8.2 & 54.2 $\pm$ 7.6 & 65.4 $\pm$ 2.9 & 78.5 $\pm$ 3.2 & 81.6 $\pm$ 12.7 & 58.5 $\pm$ 2.9 & 64.1 $\pm$ 11.4 \\
    & CheckList & 70.5 $\pm$ 3.3 & 67.1 $\pm$ 2.0 & 66.6 $\pm$ 2.6 & 65.8 $\pm$ 2.0 & 50.4 $\pm$ 16.3 & 57.8 $\pm$ 9.4 & 66.3 $\pm$ 5.1 & 81.8 $\pm$ 2.3 & 90.9 $\pm$ 1.2 & 58.0 $\pm$ 4.0 & 67.5 $\pm$ 12.9 \\
    & Semantic & 69.2 $\pm$ 3.9 & 64.3 $\pm$ 2.6 & 64.5 $\pm$ 2.5 & 64.1 $\pm$ 2.8 & 56.4 $\pm$ 16.4 & 57.3 $\pm$ 8.1 & 75.0 $\pm$ 5.8 & 78.8 $\pm$ 5.6 & 91.4 $\pm$ 1.4 & 55.9 $\pm$ 2.7 & 67.7 $\pm$ 12.6 \\
    \bottomrule
    \end{tabular}
    \caption{Average accuracy and standard deviation of \MODEL trained with different contrastive loss weighting.}
    \label{tab:lambda_detailed_accuracy}
\end{table*}

For the results of models trained with different contrastive loss weighting, refer to \Cref{tab:lambda_detailed_accuracy}.

\section{GLUE Datasets Category}
\label{section/appendix/glue_task_category}

\begin{table}
    \centering
    \small
    \begin{tabular}{c|c}
    \toprule
        Task Category & Datasets \\
        \midrule
        Sentiment Analysis & SST-2 \\
        Grammar Correctness & CoLA \\
        Paraphrase Identification & QQP, MRPC \\
        Natural Language Inference & MNLI, QNLI, RTE, WNLI \\
    \bottomrule
    \end{tabular}
    \caption{Task categories for GLUE datasets following the categories defined in PromptBench benchmark \cite{schulman_proximal_2017}.}
    \label{tab:glue_task_category}
\end{table}

Following the task category defined in PromptBench benchmark, we split the GLUE datasets into four categories as shown in \Cref{tab:glue_task_category}.

\end{document}